\documentclass[conference]{IEEEtran}
\IEEEoverridecommandlockouts
\usepackage{cite}
\usepackage{amsmath,amssymb,amsfonts}
\usepackage{algorithmic}
\usepackage{graphicx}
\usepackage{textcomp}
\usepackage{xcolor}
\usepackage{float}
\usepackage{subfig}
\usepackage{enumitem}
\def\BibTeX{{\rm B\kern-.05em{\sc i\kern-.025em b}\kern-.08em
    T\kern-.1667em\lower.7ex\hbox{E}\kern-.125emX}}
\begin{document}

\title{DQ3D: Depth-guided Query for Transformer-Based 3D Object Detection in Traffic Scenarios}

% \author{
% \IEEEauthorblockN{Ziyu Wang}
% \IEEEauthorblockA{\textit{The School of Computer Science and Engineering} \\
% \textit{BUAA}\\
% Beijing, China \\
% \and
% \IEEEauthorblockN{Wenhao Li}
% \IEEEauthorblockA{\textit{The School of Computer Science and Engineering} \\
% \textit{The University of Sydney}\\
% Sydney, Australia \\
% \and
% \IEEEauthorblockN{Ji Wu}
% \IEEEauthorblockA{\textit{The School of Computer Science and Engineering} \\
% \textit{BUAA}\\
% Beijing, China \\
\author{
\IEEEauthorblockN{
Ziyu Wang\textsuperscript{1},
Wenhao Li\textsuperscript{2},
Ji Wu\textsuperscript{1}
}
\IEEEauthorblockA{
\textsuperscript{1}The School of Computer Science and Engineering, BUAA, Beijing, China\\
}
\IEEEauthorblockA{
\textsuperscript{2}The School of Computer Science and Engineering, The University of Sydney, Sydney, Australia \\
}
\IEEEauthorblockA{
\{zoewang, wuji, wenhaol\}@buaa.edu.cn
}
}
\maketitle

\begin{abstract}
3D object detection from multi-view images in traffic scenarios has garnered significant attention in recent years. Many existing approaches rely on object queries that are generated from 3D reference points to localize objects. However, a limitation of these methods is that some reference points are often far from the target object, which can lead to false positive detections. In this paper, we propose a depth-guided query generator for 3D object detection (DQ3D) that leverages depth information and 2D detections to ensure that reference points are sampled from the surface or interior of the object. Furthermore, to address partially occluded objects in current frame, we introduce a hybrid attention mechanism that fuses historical detection results with depth-guided queries, thereby forming hybrid queries. Evaluation on the nuScenes dataset demonstrates that our method outperforms the baseline by 6.3\% in terms of mean Average Precision (mAP) and 4.3\% in the NuScenes Detection Score (NDS).
\end{abstract}


\section{Introduction}

As the complexity of urban traffic environments increasing, achieving precise road object detection has become a foundational and core technical problem.
Notably, multi-camera-based 3D object detection has garnered significant attention in both academic and industrial research, owing to its cost efficiency in comparison to LiDAR-based solutions, which generally rely on expensive sensors. 

Current multi-view 3D object detection methods can be broadly classified into two categories based on how they fuse features: dense 3D methods and sparse query-based methods. Dense 3D approaches, such as those utilizing Bird's-Eye-View (BEV) feature space \cite{li2023bevdepth,li2022bevformer,huang2022bevdet4d,huang2021bevdet} or voxel feature space \cite{li2022unifying}, render multi-view features into 3D space. 
However, these methods face scalability issues as their computational costs increase quadratically with the size of the 3D space, limiting their applicability to large-scale scenarios \cite{fan2022fully}. 
In contrast, query-based methods \cite{liu2022petr,liu2023petrv2,wang2022detr3d} employ learnable 3D object queries to aggregate multi-view image features and predict object.

\begin{figure}
    \centering
    \includegraphics[width=0.8\linewidth]{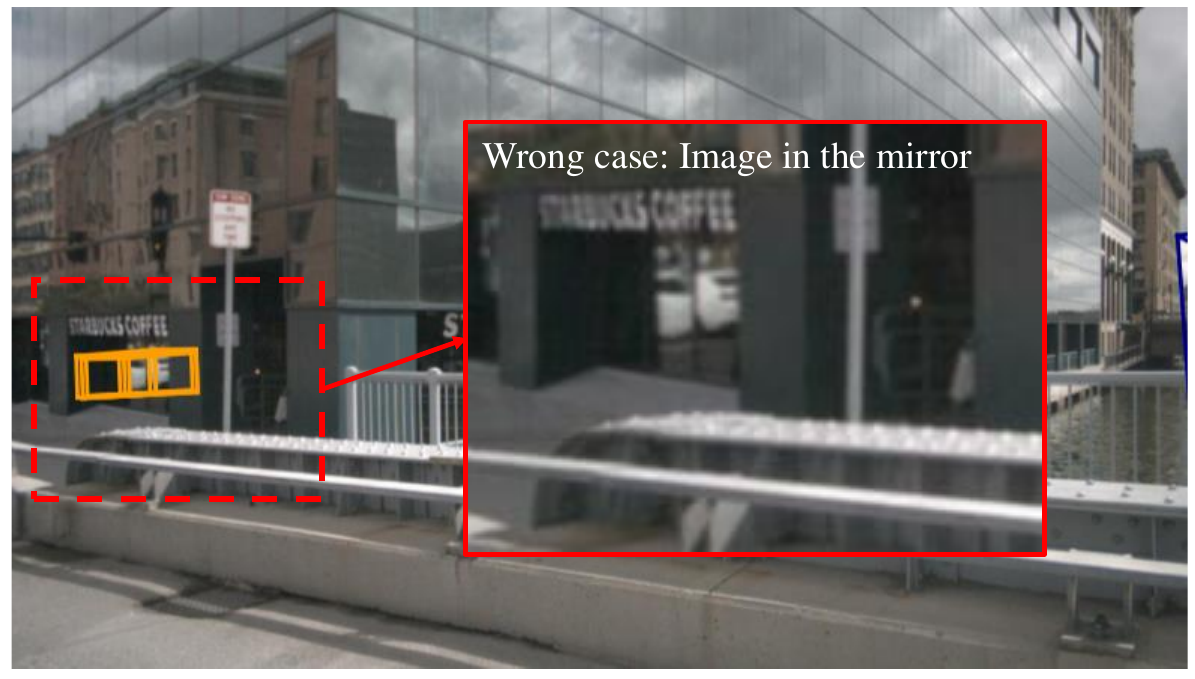}
    \caption{Wrong detection by StreamPETR. Yellow 3D box represents a detected vehicle, which is actually an image in the mirror}
    \label{fig:result example}
    \vspace{-0.5cm}
\end{figure}

Although these methods use sparse 3D object queries to mitigate computational complexity, the fixed queries are often positioned in empty space, resulting in computational inefficiency and potential false positives.
To address this issue, StreamPETR \cite{wang2023exploring} exploits the temporal consistency between two adjacent frames. It utilizes historical detection results to generate 3D object queries, which are termed temporal queries.
However, for newly appeared objects, StreamPETR still relies on fixed queries, and thus the issue of 3D object query locations being far from the object remains unresolved.
For example, as shown in Fig. \ref{fig:result example}, StreamPETR incorrectly identified the vehicle in the mirror.

\begin{figure}[t]

    \centering
    
    \subfloat[Fixed query]{\includegraphics[width=0.7\linewidth]{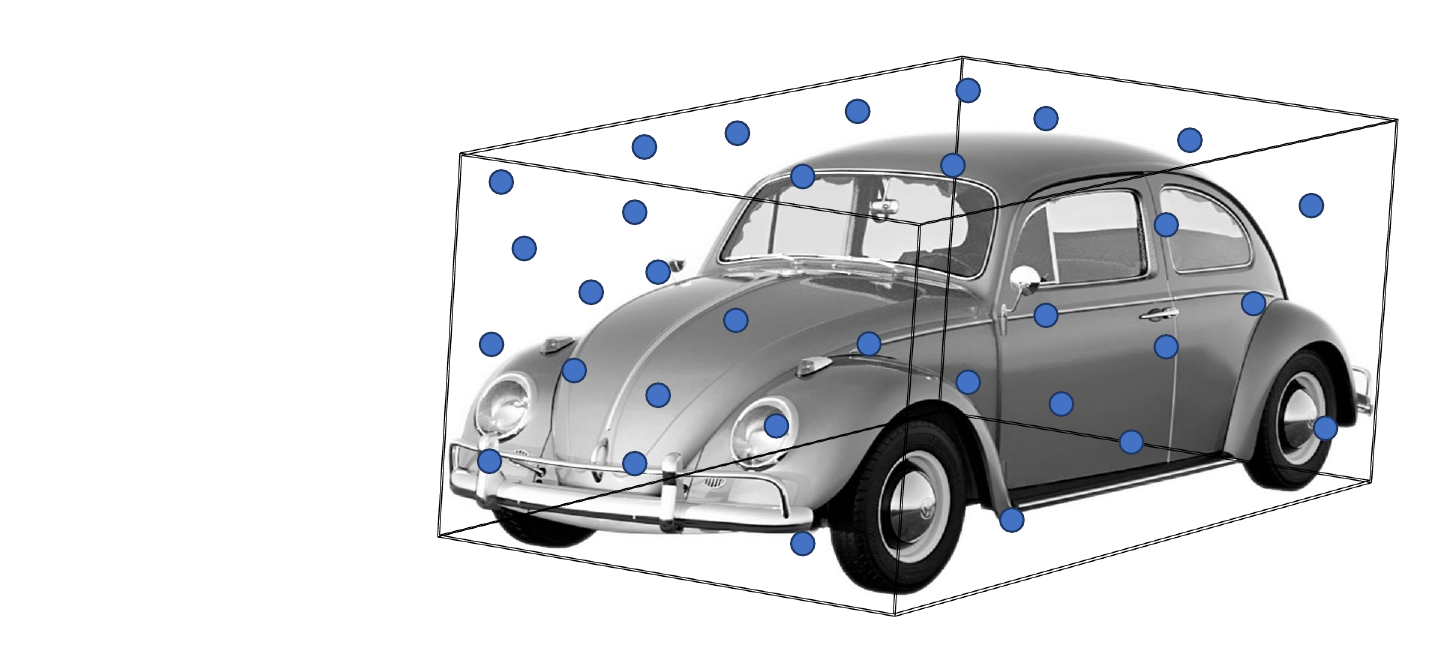}}
    \hfill
    \subfloat[Depth-guided query]{\includegraphics[width=0.7\linewidth]{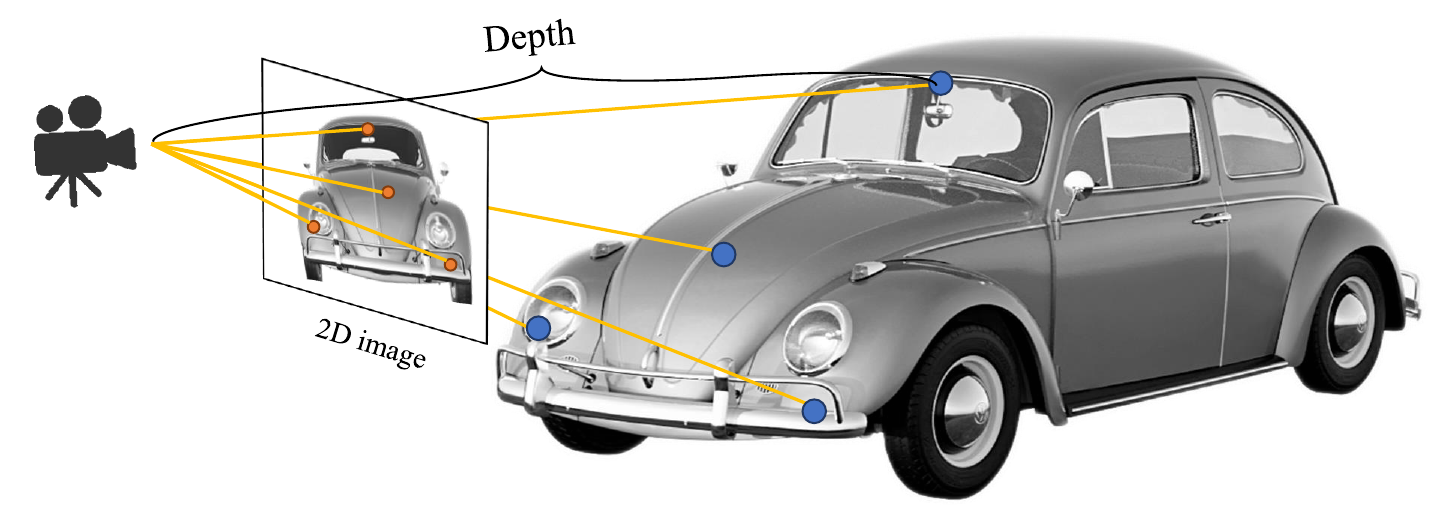}}
    \hfill

    \caption{
    An illustration of (a) the Fixed query and (b) the proposed Depth-guided query.
    The Fixed query is generated by the reference points distributed throughout 3D space, which is coarse-grained and can be distant from the object in some cases. 
    In contrast, the depth-guided query provides more accurate positional information by encoding the location of sampled points on an object's surface or within its interior, using an estimated depth.
}
\vspace{-0.5cm}
    \label{fig:query}
\end{figure}

In this paper, we introduce a simple yet effective approach that utilizes depth estimation for query generation to address the issue of 3D object query localization, as Fig.\ref{fig:query} shows.
We employ an auxiliary network to perform depth prediction to ensure that the query's reference point is located on the surface of the object or inside the object.
Moreover, to further reduce the number of reference points, we use 2D detection to constrain the region of reference points.
Additionally, inspired by StreamPETR\cite{wang2023object}, we use a temporal query alignment module to process historical detection results as temporal query to capture missing information in the current frame, such as details about partially occluded objects.
We also introduce a hybrid attention layer to fuse temporal query and the depth-based query, resulting in hybrid query. This allows historical information to be incorporated into the current frame's queries and prevents the query count from increasing with the time.
Our contributions can be summarized as: 
\begin{itemize}[leftmargin=0.3cm, itemindent=0cm]
    \item We propose a depth-guided 3D object query generation framework, DQ3D.
    This approach ensures that the generated queries are located on the surface of the object or inside the object, 
    significantly reducing the likelihood of queries focusing on irrelevant or empty areas.
    \item We introduce a hybrid attention layer to fuse historical detection result with depth-guided queries, forming hybrid queries. 
    This enables the queries to capture information about partially occluded objects in the current frame, thereby improving 3D detection performance.
    \item We evaluate DQ3D on the widely used nuScenes dataset and demonstrate significant improvements over the baseline StreamPETR under the same experimental settings. Specifically, DQ3D outperforms StreamPETR by 6.3\% in terms of mAP and 4.3\% in NDS.
\end{itemize}
.

\section{Related Work}

\subsection{Transformer-based object detection}
Object detection has been a central research topic in computer vision for decades, with traditional approaches \cite{girshick2014rich,gao2021fast,ren2016faster,ge2021yolox,redmon2016you} making significant strides in recent years. However, these methods often rely on manually designed components, such as non-maximum suppression (NMS) and anchor generation.
DETR \cite{carion2020end} represents a groundbreaking approach by framing object detection as a set prediction problem, leveraging a transformer-based architecture. This eliminates the need for heuristic target assignment and post-processing steps like non-maximum suppression. Deformable DETR \cite{zhu2020deformable} further enhances DETR by introducing deformable attention and multi-level image features, addressing issues related to slow convergence and suboptimal performance on small objects. Additionally, approaches like \cite{liu2022dab,wang2022anchor} employ anchor points or anchor boxes to provide explicit positional priors when generating object queries. Conditional DETR \cite{meng2021conditional} strengthen the cross-attention mechanisms by incorporating spatial information within the decoder embedding. 

\subsection{Camera-based 3D object detection}

Current multi-view 3D object detection methods can be broadly classified into two categories based on how they fuse features: dense 3D methods and sparse query-based methods.
Dense 3D approaches attempt to lift 2D image into 3D space, then conduct detection based on the 3D representations. BEVFormer \cite{li2022bevformer} leverages dense BEV queries to project and aggregate features from multi-view images by deformable attention . PolarFormer\cite{jiang2023polarformer} introduces the Polar representation to model BEV space. BEVDet, BEVDet4D and BEVDepth \cite{huang2022bevdet4d,li2023bevdepth,huang2021bevdet} adopt the Lift-Splat module \cite{philion2020lift} to transform multi-view image features into the BEV representation based on the predicted depth distribution.
% Although such 3D space representations are conducive to unifying multi-view images, the memory consumption and computational cost would increase with the enlarging of the detection range in 3D space. 
Another line of research utilizes learnable 3D object queries to aggregate image features and predict objects, following the principles of DETR \cite{carion2020end}. These methods, known as sparse query-based approaches, include DETR3D \cite{wang2022detr3d}, which generates 3D reference points from object queries and projects them onto multi-camera images using camera parameters. 
% The query features are then refined by sampling corresponding point features from the images.
In contrast to establishing a fixed 3D-to-2D query mapping, some approaches introduce more flexible mappings via attention mechanisms. The PETR series \cite{liu2022petr,liu2023petrv2,wang2023focal,shu20233dppe,wang2022detr3d} incorporates 3D position-aware image features and learns a flexible mapping between queries and image features through global cross-attention.
Building on the frame-to-frame consistency in video streams, StreamPETR \cite{wang2023object} introduces historical detection results as temporal queries to leverage past information and enhancing the performance of the PETR.
% Nonetheless, in the absence of guidance, these approaches generally rely on a substantial number of 3D object queries distributed throughout 3D space to achieve sufficient object recall.

% \subsection{2D Auxiliary Tasks for 3D Detection}
% 3D object detection from surround-view images has the potential to benefit from 2D auxiliary tasks, with various studies \cite{xie2022m, zhang2023simple, wang2023focal} investigating approaches to leverage this potential. These methods generally encompass strategies such as 2D pretraining, auxiliary supervision, and proposal generation. For example, SimMOD \cite{zhang2023simple} utilizes sample-wise object proposals and introduces a two-stage training framework, wherein perspective object proposals are first generated and then refined through iterative steps in a manner similar to DETR3D. In contrast, Focal-PETR \cite{wang2023focal} incorporates 2D object supervision to adaptively focus the attention of 3D queries on key foreground regions. BEVFormerV2 \cite{yang2023bevformer} proposes a two-stage BEV detection pipeline, in which perspective proposals are passed to the BEV head for final prediction.

\section{Method}
\begin{figure*}[t]
    \centering
    \includegraphics[width=0.7\linewidth]{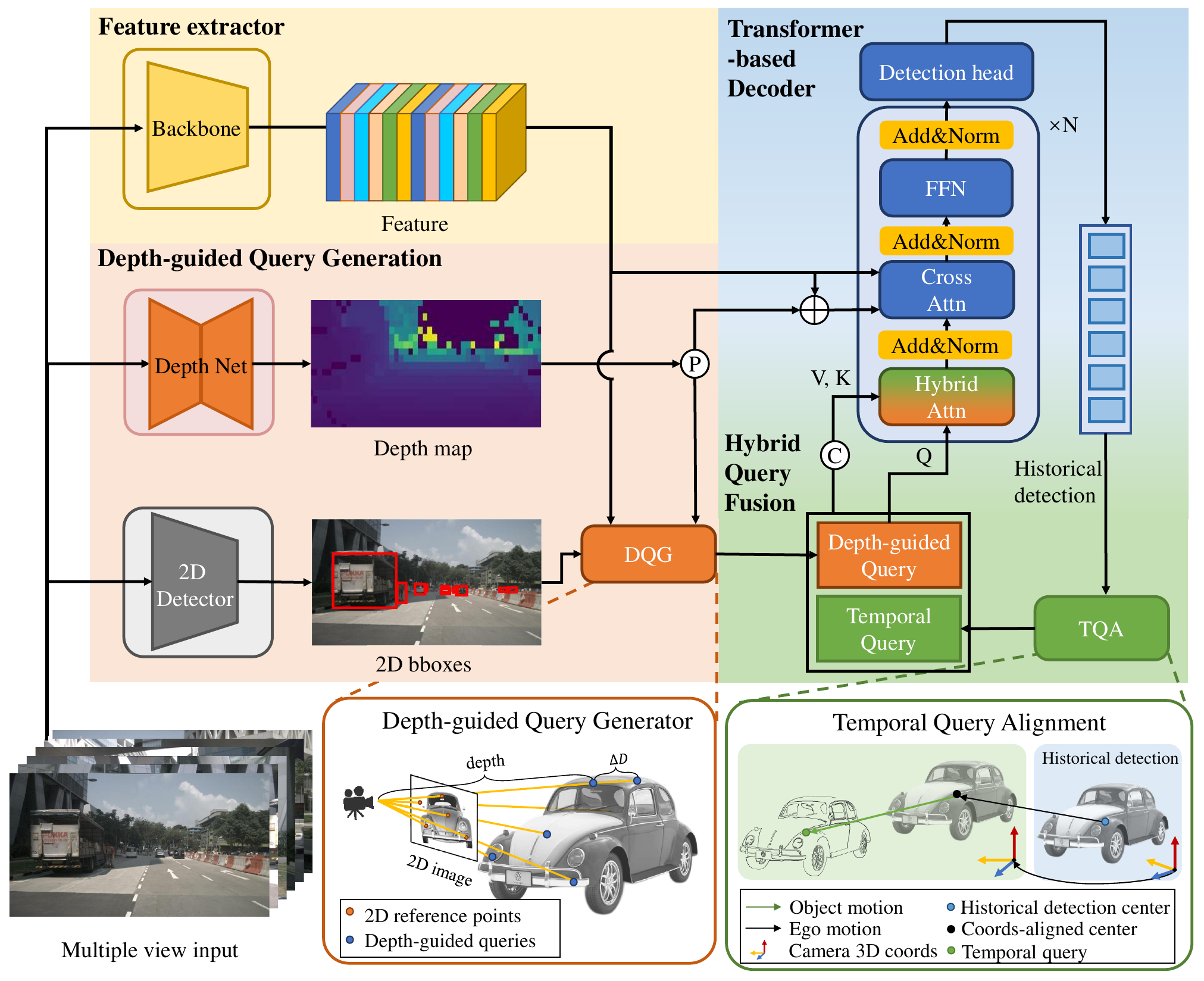}
    \caption{
    The proposed DQ3D framework. We introduce two key innovations: the Depth-guided Query Generator (DQG) that initializes depth-guided queries using depth maps and 2D detection boxes, and a hybrid attention layer in the decoder to fuse depth-guided and temporal queries for enhanced 3D object detection.
    % The framework of the proposed DQ3D. 
    % % Given the input multi-view images, image feature maps are extracted by a feature extractor. Meanwhile, a 2D detector and a depth net are used to obtain per-view 2D detection results and depth map. 
    % We use two auxiliary networks, depth net and 2D detector, to extract depth map and 2D detection boxes from given multi-view images.  
    % Depth-guided query generator(DQG) takes object features, 2D detection boxes and depth map as input to initialize depth-guided queries. 
    % % Position Encoder is used to embed 3D position into the query and the image feature.
    % Historical detection result is used as temporal query to capture missing information in the current frame inspired by StreamPETR\cite{wang2023object}.
    % We introduce hybrid attention layer in decoder to fuse depth-guided query and temporal query together. 
    % % Lastly, a detection head is applied to generate 3D detection results.
    }\vspace{-0.5cm}
    \label{fig:structure}
\end{figure*}
\subsection{Overview}
In this section, we will introduce our method DQ3D in detail as shown in Fig.\ref{fig:structure}. 
Following transformer-based detection methods\cite{wang2022detr3d,liu2022petr,liu2023petrv2,shu20233dppe,carion2020end}, our architecture includes a backbone network $\mathcal{N}_B$ to extract feature from multi-camera images,
a position encoder $PE_{3D}$ to embed 3D position into image feature
, a decoder transformer $\mathcal{N}_D$ 
% to process the interaction between the feature and a set of sparse object queries
, and a detection head that makes the final detection prediction.
Three additional components are introduced in our method, which are described below:
a depth network $\mathcal{N}_{dep}$, a 2D detector $\mathcal{N}_{2D}$ and a depth-guided query generator (DQG) that uses the depth map and 2D detections as inputs to generate queries.
More over, historical detection result is used as temporal query to capture missing information in the current frame inspired by StreamPETR\cite{wang2023object}.
We introduce hybrid attention layer as a replacement of self-attention in decoder to fuse depth-guided query and temporal query together. 
\subsection{Preliminary}
\begin{figure*}[t]
    \centering
    \includegraphics[width=0.7\linewidth]{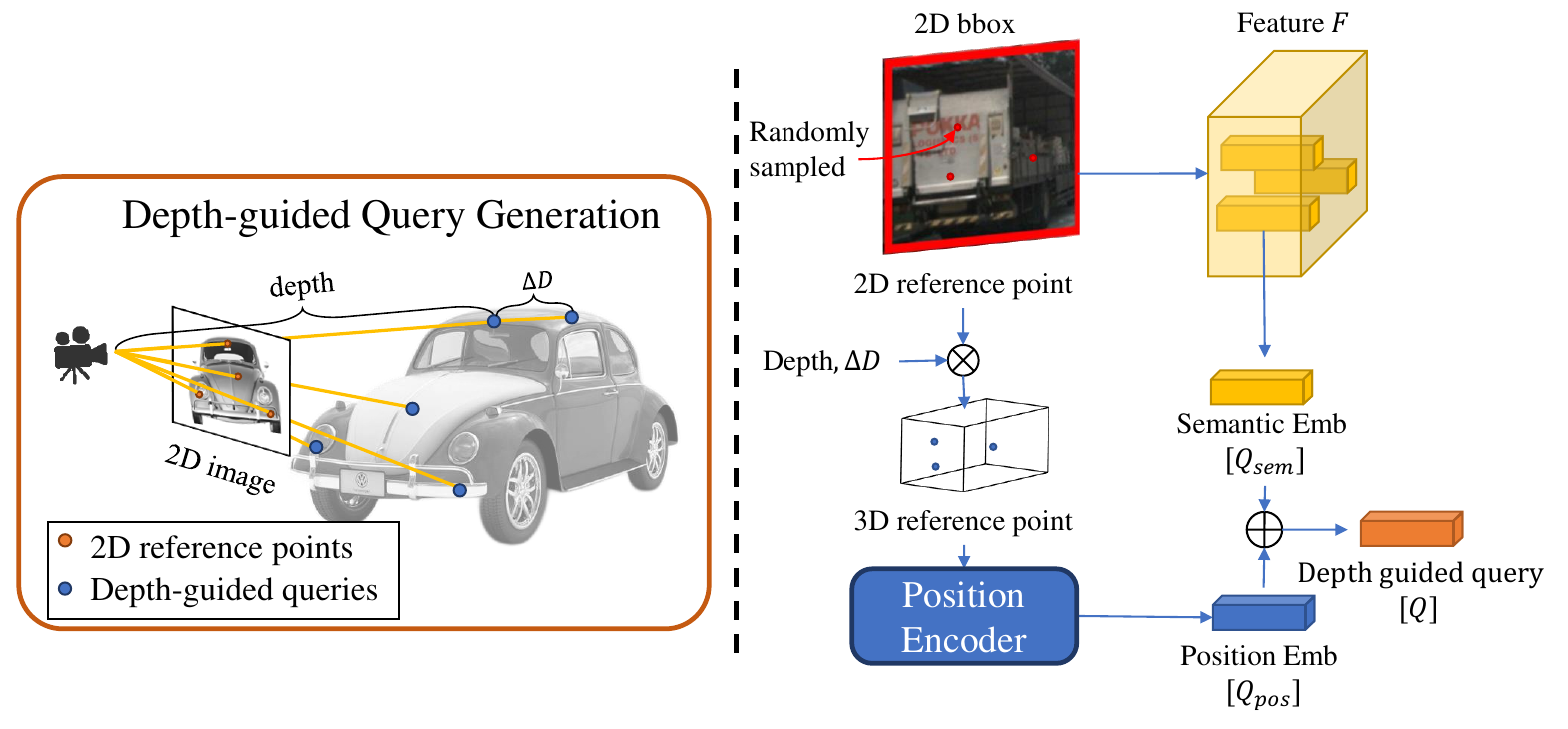}
    \caption{
    Depth-guided query generator(DQG). DQG sample the 3D reference points using 2D detection box and depth map $D_{pred}$.    3D reference points are not only sampled on the surface of objects, but also deeper within the object based on the interval $\Delta D$. 
    The position embedding of the query is computed based on the 3D coordinates of the reference point.
    The query's semantic embedding is sampled from image feature according to the 2D reference point position.}
    \label{fig:depth-guided}
    \vspace{-0.5cm}
\end{figure*}
In this section, we introduce some foundational concepts and methods that are essential for understanding our proposed method, including 2D to 3D transformation and 3D point position encoder.

\subsubsection{2D to 3D Transformation}
2D to 3D Transformation is a basic method for our method to locate reference's 3D position from 2d position, depth and camera parameters.
% to transform 2D  position with depth and camera parameters into 3D position. 
Assuming we have estimated the depth$D_{pred}$ for surrounding-view image $x$, which is denoted as follows.
\begin{equation}
\label{2D to 3D}
\begin{aligned}
        &D_{pred} \gets \mathcal{N}_{dep}(x)\\
    &P_{3D} \gets Trans_{2D \to 3D}(P_{2D},D_{pred})
\end{aligned}
\end{equation}
Here, $P_{3D}$ and $P_{2D}$ refer to the 3D position $(x, y, z)$ and 2D position $(u, v)$ respectively. 
In detail, we take a pixel position $(u, v)$ in the view of the $i^{th}$ camera, and sample the predicted depth $d = D_{pred}[:, u, v]$ as input, then output its 3D coordinate $(x, y, z)$ with the equation below:
\begin{equation}
\begin{aligned}
    P_{3D}[:,u,v]=
    \left[
    \begin{array}{c}
    x \\y \\z \\1
    \end{array}
    \right]\gets R_iK_i^{-1}    \left[
    \begin{array}{c}
    ud \\vd \\d \\1
    \end{array}
    \right]+T_i
\end{aligned}
\end{equation}
Here, 3D coordinate system is based on ego position and pose. $R_i, T_i \in \mathbb{R}^{4 \times 4}$ are the extrinsic parameters of the $i$-th camera, where $R_i$ is the rotation matrix and $T_i$ is the transformation matrix, both defined relative to the ego pose. The intrinsic matrix $K_i \in \mathbb{R}^{4 \times 4}$ of the $i$-th camera is computed as follows.
\begin{equation}
\begin{aligned}
    K = \left(
    \begin{array}{cccc}
    f_x&0&c_x&0\\
    0&f_y&c_y&0\\
    0&0&1&0\\
    0&0&0&1
    \end{array}
\right)    
\end{aligned}
\end{equation}
Here, $f_x$ and $f_y$ denote the focal length of this camera, adjusted for width and height scaling, and $c_x,c_y$ represent the center position of the image, adjusted for image cropping. 

\subsubsection{3D Point Position encoder}
Position encoding is a crucial technique used in transformer-based models to inject spatial information into the model, allowing it to understand the relative or absolute positions of elements in a sequence.
We adopt 3D point position encoder (3DPPE\cite{shu20233dppe}) to encode 3D positional information as 3D point position embedding, as Eq.\ref{Eq:3dppe} shows.
\begin{equation}
\label{Eq:3dppe}
    \begin{aligned}
        PE_{3D}[:,u,v] = MLP(Cat(
        &Sine(P_{3D}[0,u,v]),\\
        &Sine(P_{3D}[1,u,v]),\\
        &Sine(P_{3D}[2,u,v])))
    \end{aligned}
\end{equation}
where the sine/cosine positional encoding function $Sine$ maps a 1-dimensional coordinate value to a $\frac{C}{2}$-dimensional vector used in \cite{vaswani2017attention}, the sequential $Cat$ operator concatenate $Sine(P_{3D}[0,u,v])$, $Sine(P_{3D}[1,u,v])$ and $Sine(P_{3D}[2,u,v])$ to generate a $\frac{3C}{2}$-dimensional vector, then the $MLP$ consisted of two linear layrs and a RELU activation reduces the vector dimension from $\frac{3C}{2}$ to $C$.

\subsection{Depth-guided Query Generator}

The aim of Depth-guided Query Generator (DQG) is to use depth map $D_{pred}$ and 2D detection result $bbox_{2D}$ to generate object queries $Q$ which is localized near the surface of 3D objects,
as shown in Figure\ref{fig:depth-guided}. 
 % Subsequently, we compute two types of embeddings for each reference point: the position embedding \( Q_{pos} \), which is derived from 3D coordinates of the reference point, and the semantic embedding \( Q_{sem} \), which is obtained by incorporating 2D position of reference point and image feature. Finally, the query is constructed as the combination of these two embeddings, as:
% \begin{equation}
    % Q \gets Q_{pos}+Q_{sem}
% \end{equation}

Assuming that we have detected 2d bounding boxes $bbox_{2D}$ and depth map $D_{pred}$ from surrounding view using the 2D detector $\mathcal{N}_{2D}$ and depth net$\mathcal{N}_{dep}$,
% We introduce a Depth-guided query generator module to sample 2D reference points from 2D detection result.
DQG first samples several reference points based on the 2D detection results and then project them into 3D space using the depth map. 
Considering the potential for occlusions within 2D bounding boxes, we do not rely solely on the center point of the 2D bounding box as the reference point, instead, additional points are randomly sampled within the bounding box for better representation. 
The sampled 2D reference points are transformed into 3D reference points guided by the depth map. To better align with the 3D object, we not only sample 3D reference points on the object's surface but also sample additional reference points deeper within the object.
\begin{equation}
\begin{aligned}
    &bbox_{2D} \gets \mathcal{N}_{2D}(x)\\
    &P_{2D} \gets sample(bbox_{2D})\\
    &P_{ref} \gets Trans_{2D \to 3D}(P_{2d},D_{pred}+i*\Delta D), i=0,1,...\\
\end{aligned}
\end{equation}
Here, \( Trans_{2D \to 3D} \) refers to the transformation introduced in Eq. \ref{2D to 3D}, which maps 2D pixel coordinates to their corresponding 3D coordinates. \( D_{pred} \) denotes the predicted depth map, and \( P_{ref} \) represents the 3D reference point we sample based on the predicted depth. \( \Delta D \) is a predefined sampling interval that helps in selecting points along the depth direction. 

After fetching 3D reference points, DQG compute two types of embeddings for each reference point: the position embedding \( Q_{pos} \), which is derived from 3D coordinates of the reference point using Eq. \ref{Eq:3dppe}, and the semantic embedding \( Q_{sem} \), which  is derived from the image feature \( F \) at 2D position \( (u, v) \) in the view of the \( i \)-th camera.  In a summary,
the depth-guided object query is computed as follows.

\begin{equation}
\begin{aligned}
    &Q_{pos} \gets PE_{3D}(Norm(P_{ref}))\\
     &   Q_{sem} \gets F[i,:,P_{2D}]\\
    &Q \gets Q_{pos}+Q_{sem}\\
\end{aligned}
\end{equation}

\begin{figure*}[t]
    \centering
    \includegraphics[width=0.8\linewidth]{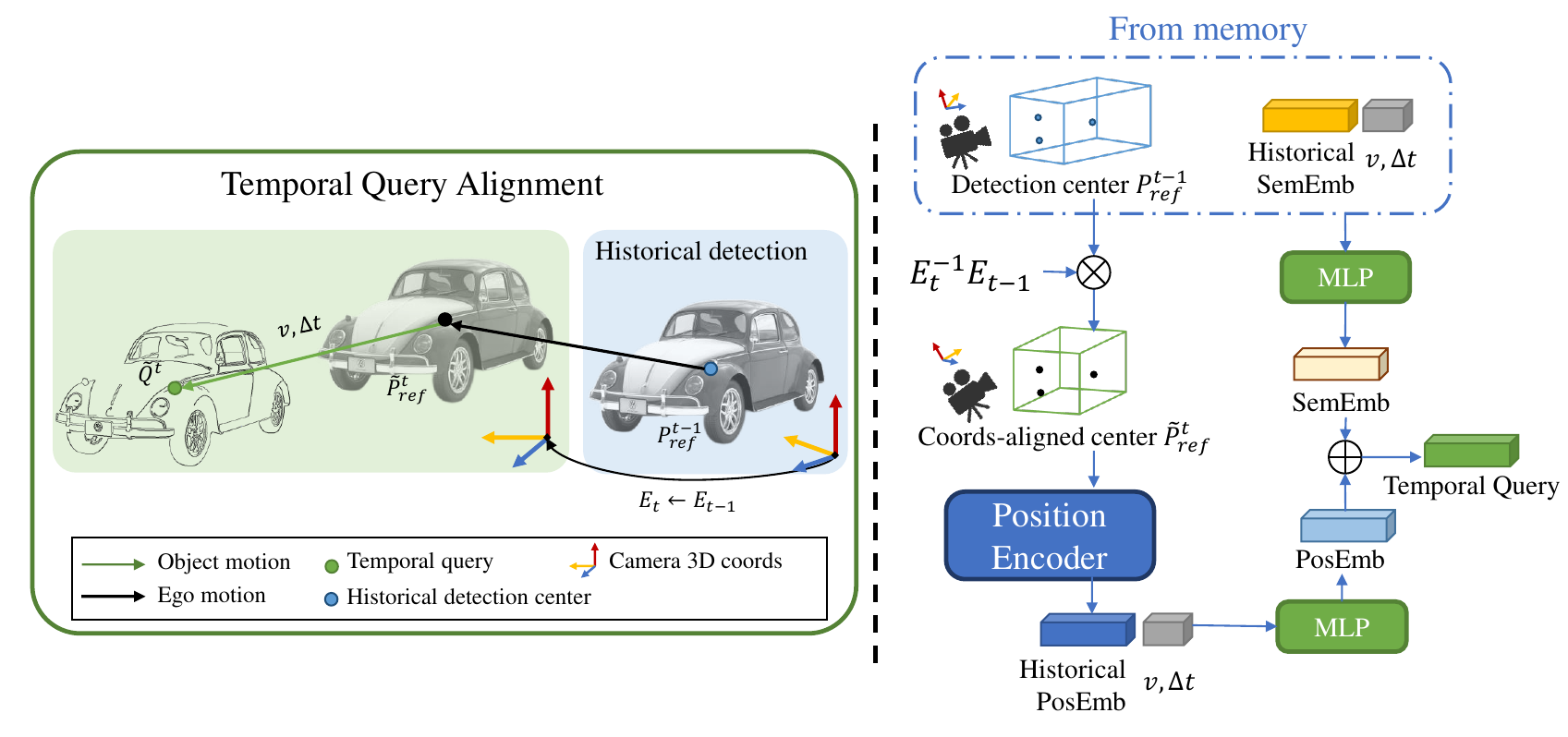}
    \caption{Temporal query alignment(TQA). The center point$P_{ref}^{t-1}$ and semantic embedding $Q_{sem}^{t-1}$ of historical detection result is saved. TQA align these information with current frame by coordinate transform and MLP considering the ego motion and the object motion.}
    \label{fig:tempor}
    \vspace{-0.5cm}
\end{figure*}
\subsection{Temporal Query Alignment}

Inspired by StreamPETR\cite{wang2023exploring}, 
we also use historical detection result for query generation, which is named Temporal query. 
A memory queue of $N \times K$ is used for effective temporal modeling.
Here $N$ is the number of stored frames and $K$ is the number of objects stored per frame.
For each selected object, we save the relative time interval $\Delta t$, query's semantic embedding $Q^t_{sem}$, reference center point position $P^t_{ref}$, object velocity $v$, and ego-pose matrix $E$ are stored in memory queue.

% To leverage the benefits of both types of queries, we introduce a Temporal Query Alignment (TQA) module 
% and a hybrid attention layer to fuse temporal queries and depth-guided queries into hybrid queries, which will be explained in detail below.

TQA (Temporal Query Alignment) is designed to align the historical queries to current frame by modeling the movement of objects and ego camera. For simplicity, we take the transformation process from the last frame $t-1$ as the example and adopt the same operation for other previous frames, as shown in Fig.\ref{fig:tempor}.
Considering the motion of ego camera $E^{t-1} \to E^t$, we initially compute position embedding after aligning the coordinates of different frames:
\begin{equation}
\begin{aligned}
    &\tilde{P}^t_{ref} \gets (E^t)^{-1}E^{t-1}P^{t-1}_{ref}\\
    &\tilde{Q}_{pos}^{t} \gets PE_{3D}(Norm(\tilde{P}_{ref}^{t}))\\
\end{aligned}
\end{equation}
where $P^{t-1}_{ref}$ is the bounding box center points position of frame $t-1$, and the $E^t$ is the quaternion which denotes the pose of ego camera on frame$t$. 
Considering the motion of objects, we predict the displacement of each object by the predicted motion attributes ($v,\Delta t$).
We use two MLPs to describe the change of the position embedding and the semantic embedding, and the temporal query is the sum of both,
as equation follows:
\begin{equation}
\begin{aligned}
        &\tilde{Q}_{pos}^{t}\gets MLP(\tilde{Q}_{pos}^{t},v,\Delta t)\\
        &\tilde{Q}_{sem}^{t}\gets MLP(Q_{sem}^{t-1},v,\Delta t)\\
        &\tilde{Q}^t \gets \tilde{Q}^t_{pos}+\tilde{Q}^{t}_{sem}
\end{aligned}
\end{equation}

\subsection{Hybrid Attention Layer}
\begin{figure}[t]
    \centering
    \includegraphics[width=0.8\linewidth]{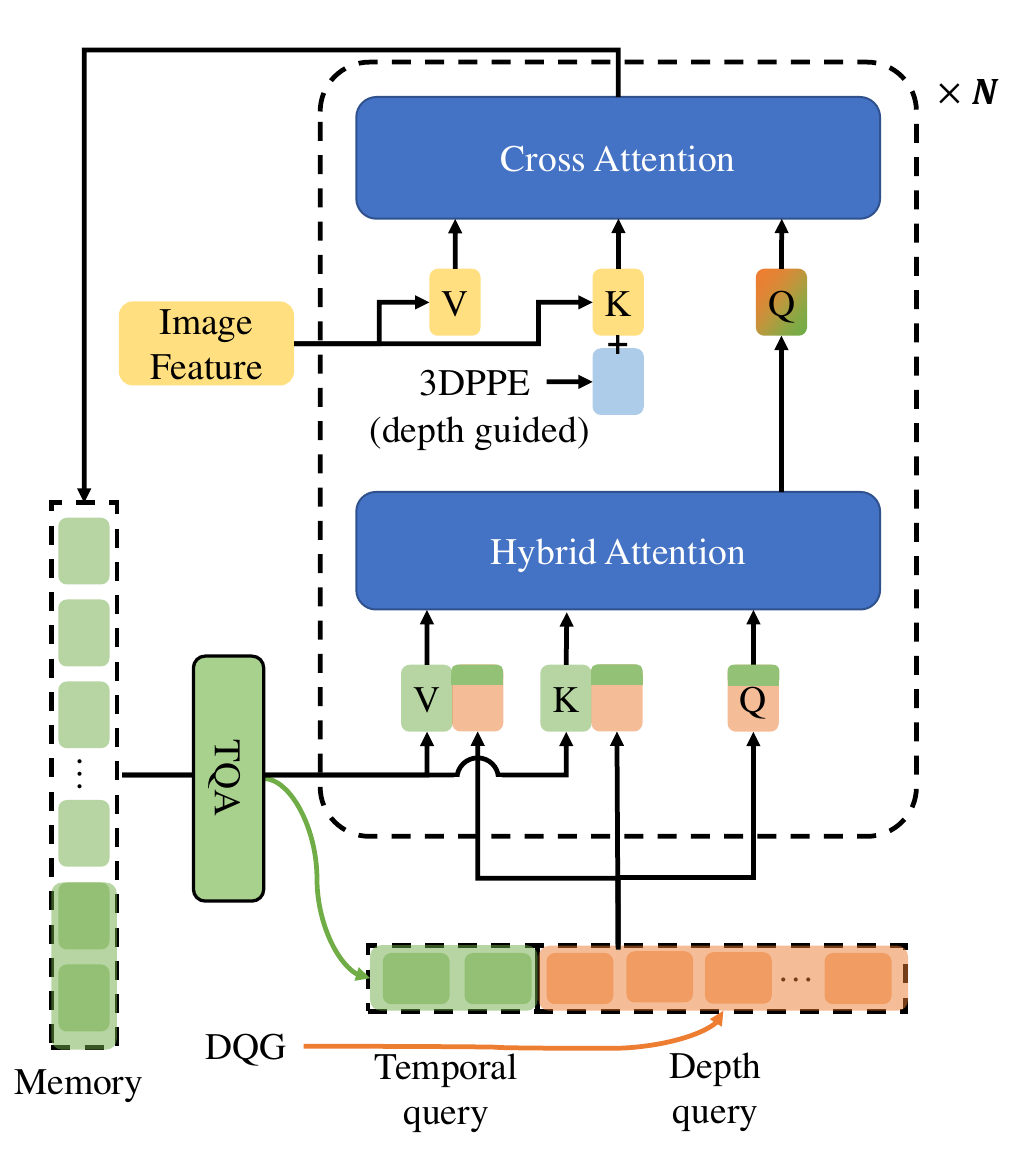}
    \caption{
    Decoder in our method.
    Compare to the traditional DETR decoder, our decoder replace self-attention layer with hybrid attention layer to fuse temporal queries and depth-guided queries,  resulting in hybrid queries for subsequent decoding.    
%     Inspired by StreamPETR\cite{wang2023object}, part of the latest temporal queries are propagated into current object queries to fully utilize the spatial and context priors in streaming video.
% Image's position information is encoded by 3D Point Position Encoder(3DPPE\cite{shu20233dppe}) in our method.
    }
    \vspace{-0.4cm}
    \label{fig:decoder}
\end{figure}

% 为什么要引入？
 We replace the self-attention layer in previous work\cite{carion2020end,liu2022petr,wang2022detr3d,shu20233dppe} with hybrid attention, which introduces temporal interaction. 
 This not only allows historical information to be incorporated into the current frame's queries but also prevents the query count from increasing with the length of time. 
 As shown in Fig. \ref{fig:decoder}, temporal queries from the memory queue and from the DQG are concatenated as the value and key matrices, 
 then the hybrid attention layer fuse them with depth-guided query, 
 resulting in hybrid queries for subsequent decoding. We denote this process as:
\begin{equation}
    \begin{aligned}
        &X= Cat(Q_{temp},Q_{dep})\\
        &Q,\ K,\ V=W_qQ_{dep},\ W_kX,\ W_qX\\
        &Q_{hybrid} = Attn(Q,K,V)= softmax\left(\frac{QK^T}{\sqrt{d_k}}V\right)
    \end{aligned}
\end{equation}
Here, \( Q_{\text{temp}} \) and \( Q_{\text{dep}} \) refer to the temporal query and depth-guided query, respectively. \( W_v \), \( W_k \), and \( W_q \) denote three learnable matrices used to compute the query (\( Q \)), key (\( K \)), and value (\( V \)), respectively. Additionally, \( d_k \) represents the normalization parameter.

\subsection{Loss Functions}
The 2D object detector and depth estimator are fixed in our method. 
For 3D object detection loss, we follow previous works \cite{wang2022detr3d,liu2022petr,liu2023petrv2} to use Hungarian algorithm \cite{kuhn1955hungarian} for label assignment. 
Focal loss $\mathcal{L}_{cls3D}$ \cite{lin2017focal} and L1 loss $\mathcal{L}_{reg3D}$ are adopted for classification and box regression respectively. The 3D object detection loss can be computed as follows.
\begin{equation}
\begin{aligned}
        &p_t = \begin{cases}
            p, & \text{if True Positive} \\
            1 - p, & \text{otherwise}
        \end{cases} \\
        &\mathcal{L}_{cls3D} = -\sum_{i=0}^{n}(1-p_t)^\gamma \log(p_t) \\
\end{aligned}
\end{equation}
\begin{equation}
% \begin{aligned}
        \mathcal{L}_{reg3D} = \sum |y_{pred} - y_{gt}| 
% \end{aligned}
\end{equation}
\begin{equation}
        \mathcal{L}_{3D} = \lambda_{cls3D} \mathcal{L}_{cls3D} + \mathcal{L}_{reg3D}
\end{equation}
Here, $p$ denotes the classification score.
$y \in$ \{position, size, velocity, orientation\} is the feature of the object. $\lambda_{cls3D}$ is the balance weight of two losses.

\section{Experiments}
\subsection{Dataset}
The nuScenes dataset\cite{caesar2020nuscenes} consists of 1000 sequences of various scenes captured in both Boston and Singapore, where each sequence is approximately 20 seconds long. The dataset is officially partitioned into training, validation, and testing subsets with 700,150, and 150 sequences, respectively. For each sample, we have access to the six surrounding cameras as well as the camera calibrations.

\subsection{Implementation Details}

\subsubsection{Network Architecture}
With StreamPETR as our baseline, 
we adopt VoVNet99 \cite{lee2019energy} pre-trained with FCOS3D \cite{wang2021fcos3d} on nuScenes as the backbone to conduct main experiments. The output of the P4 stage of VoVNetV2 is used as the image feature.
ViT-Large \cite{dosovitskiy2020image} pre-trained by Objects365 \cite{shao2019objects365} and COCO \cite{lin2014microsoft} dataset is used to scale up our model. 
The decoder contains 6 transformer decoder layers, followed by a MLP head for classification and regression.
Faster-RCNN\cite{ren2016faster} serves as 2D detector in our experiments. The 2D score threshold and Non-maximum Suppression (NMS) IoU threshold are set to 0.05 and 0.7 respectively.
We use SurroundDepth\cite{wei2023surrounddepth} as metrics depth network to estimate image-size depth map, the minimum and maximum depth are 0.05m and 80m respectively.
% The depth and object detection perception range is set as 152.4m × 152.4m.
%%%%%%%%%%%%%%%%%%%%%%%%%%%%%%%%%%%%%%%%%%
\subsubsection{Training and evaluation}
We use AdamW\cite{loshchilov2017decoupled} optimizer with a weight decay of 0.01, the total batch size 4,
cosine annealing policy \cite{loshchilov2016sgdr} with the initial learning rate 1e-4. The models are totally trained for 24 epochs, following the previous method\cite{liu2022petr,wang2023object,shu20233dppe}.
All experiments were performed on 14 vCPU Intel(R) Xeon(R) Platinum 8362 CPU @ 2.80GHz and a 48GB RAM and NVIDIA RTX 3090*2 GPUs. Our implementation is based on MMDetection3D \cite{chen2019mmdetection}.
\begin{figure}[t]
    \centering
    \includegraphics[width=\linewidth]{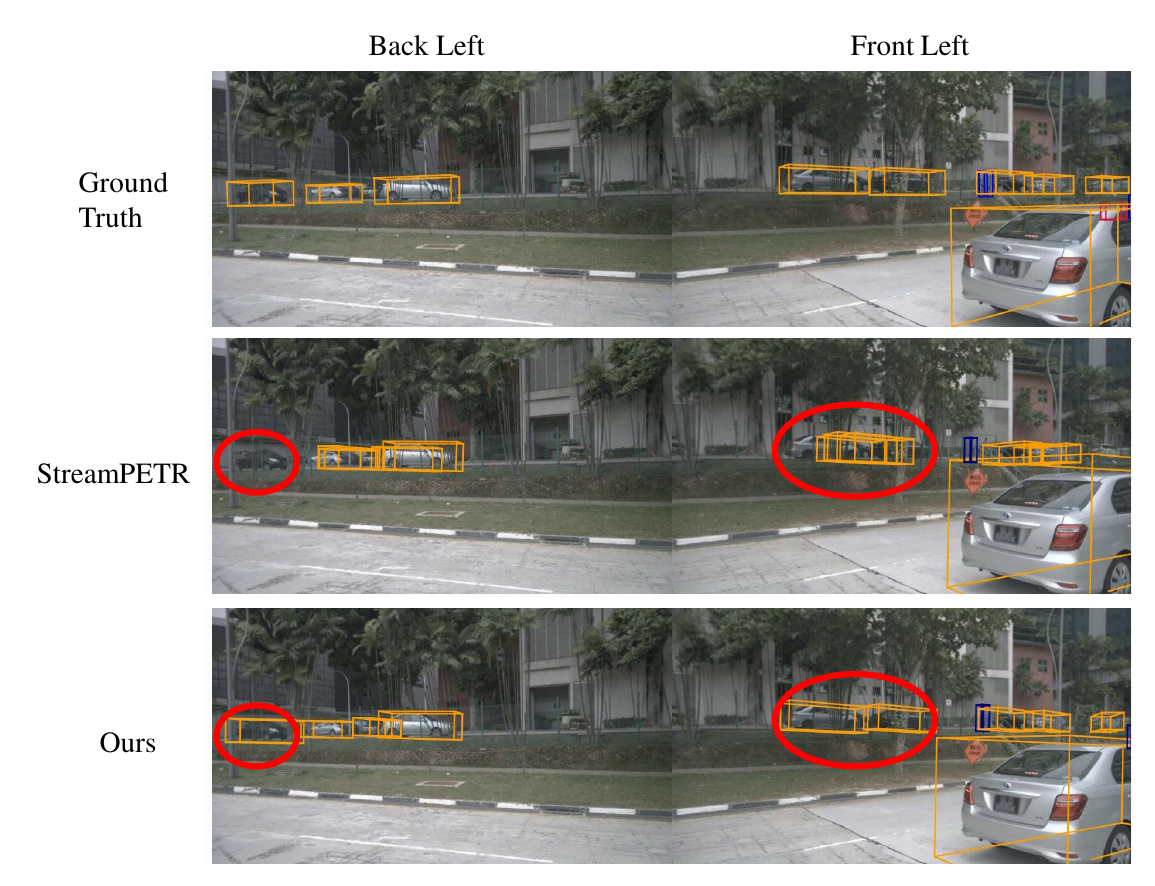}
    \caption{A case where the baseline StreamPETR method fails while our proposed DQ3D method successfully to detect the object. The red circles highlight the accurately detected targets by DQ3D. }
    \label{fig:vis}
    \vspace{-0.3cm}
\end{figure}
\begin{table}[t]
    \centering
    \tabcolsep=2.5pt
    \resizebox{0.5\textwidth}{!}{
    
    \begin{tabular}{c|c c|c c c c c}
    \hline
        Methods&mAP$\uparrow$& NDS$\uparrow$& mATE$\downarrow$&mASE$\downarrow$&mAOE$\downarrow$&mAVE$\downarrow$&mAAE$\downarrow$\\
    \hline
    FCOS3D$\dagger$\cite{wang2021fcos3d}&0.295& 0.372& 0.806& 0.268& 0.511& 1.315& \textbf{0.170}\\
    PGD$\dagger$\cite{wang2022probabilistic}   & 0.335& 0.409& 0.732& 0.263& 0.423& 1.285& 0.172\\
    DETR3D$\dagger$\cite{wang2022detr3d}   &0.349& 0.434& 0.716& 0.268& 0.379& 0.842& 0.200\\
    BEVFormer$\dagger$\cite{li2022bevformer}   & 0.375& 0.448& 0.725& 0.272& 0.391& 0.802& 0.200\\
    Ego3RT$\dagger$\cite{lu2022learning}   & 0.375& 0.450 &0.657 &0.268& 0.391& 0.850& 0.206\\
    SpatialDETR$\dagger$\cite{doll2022spatialdetr}   &0.351& 0.425 &0.772& 0.274 &0.395 &0.847 &0.217\\
     PETR$^*$\cite{liu2022petr}&0.378&0.426&0.746& 0.272& 0.488&0.906&0.212 \\
     3DPPE$^*$\cite{shu20233dppe}& 0.398&0.446&0.704&0.270&0.495&0.843&0.218\\\hline
     PETRv2$^*$\cite{liu2023petrv2}&0.410&0.503&0.723&0.263&0.453&0.389&0.193\\
     StreamPETR$^*$\cite{wang2023object}&0.482&0.571&0.610&\textbf{0.256}&\textbf{0.375}&0.263&0.194\\
     Ours$^*$&\textbf{0.498}&\textbf{0.582}&\textbf{0.585}&0.260&0.384&\textbf{0.240}&0.199\\
    \hline
    \end{tabular}
    }
    \caption{Comparison of other methods on nuScenes val set. $*$ denotes the input image size is $320 \times 800$ and backbone V2-99, and $\dagger$ denotes the input image size is $900 \times 1600$ and backbone Resnet-101. The method above the horizontal line uses a single-frame data, while the one below utilizes multi-frame data.}
    \label{tab:performance on val}
    \vspace{-0.4cm}
\end{table}
\begin{table}[t]
    \centering
    \resizebox{0.5\textwidth}{!}{
    \tabcolsep=2pt
    
    \begin{tabular}{c|c c|c c c c c}
    \hline
         Methods &mAP$\uparrow$& NDS$\uparrow$& mATE$\downarrow$&mASE$\downarrow$&mAOE$\downarrow$&mAVE$\downarrow$&mAAE$\downarrow$\\
    \hline
    DD3D$^*$\cite{park2021pseudo}&0.418&0.477&0.572&0.249&0.368&1.014&0.124\\
    DETR3D$^*$\cite{wang2022detr3d}&0.412&0.479&0.641&0.255&0.394&0.845&0.133\\
    Ego3RT$^*$ \cite{lu2022learning}&0.425&0.473&0.549&0.264&0.433&1.014&0.145\\
    BEVDet$^*$ \cite{huang2021bevdet}&0.424&0.488&0.524&0.242&0.373&0.950&0.148\\
    BEVFormer$^*$\cite{li2022bevformer}&0.435&0.495&0.589&0.254&0.402&0.842&0.131\\
    SpatialDETR$^*$\cite{doll2022spatialdetr}&0.424&0.486&0.613&0.253&0.402&0.857&0.131\\
     PETR$^*$\cite{liu2022petr}&0.441&0.504&0.593& 0.249& 0.383&0.808&0.132 \\
     3DPPE$^*$\cite{shu20233dppe}& 0.460&0.514&0.569&0.225&0.394&0.796&0.138\\\hline
     PETRv2$^\#$\cite{liu2023petrv2}&0.490&0.582&0.561&0.243&0.361&0.343&\textbf{0.120}\\
     StreamPETR$^\#$\cite{wang2023object}&0.550&\textbf{0.631}&0.493&\textbf{0.241}&\textbf{0.343}&\textbf{0.243}&0.123\\
     StreamPETR$^*$&0.503&0.582&0.555&0.253&0.456&0.307&0.124\\
     Ours$^*$&\textbf{0.566}&0.625&\textbf{0.448}&0.259&0.464&0.283&0.132\\
    \hline
    \end{tabular}
    }
    \caption{Comparison of other methods on nuScenes test set.$*$ denotes the input image size is $320 \times 800$, and $\#$ denotes the input image size is $640 \times 1600$. The method above the horizontal line uses a single-frame data, while the one below utilizes multi-frame data. All models are train on V2-99 backbone.}
    \label{tab:performance on test}
    \vspace{-0.5cm}
\end{table}

\subsection{Performance of DQ3D}

\begin{figure*}[t]
    \centering
    \includegraphics[width=0.85\linewidth]{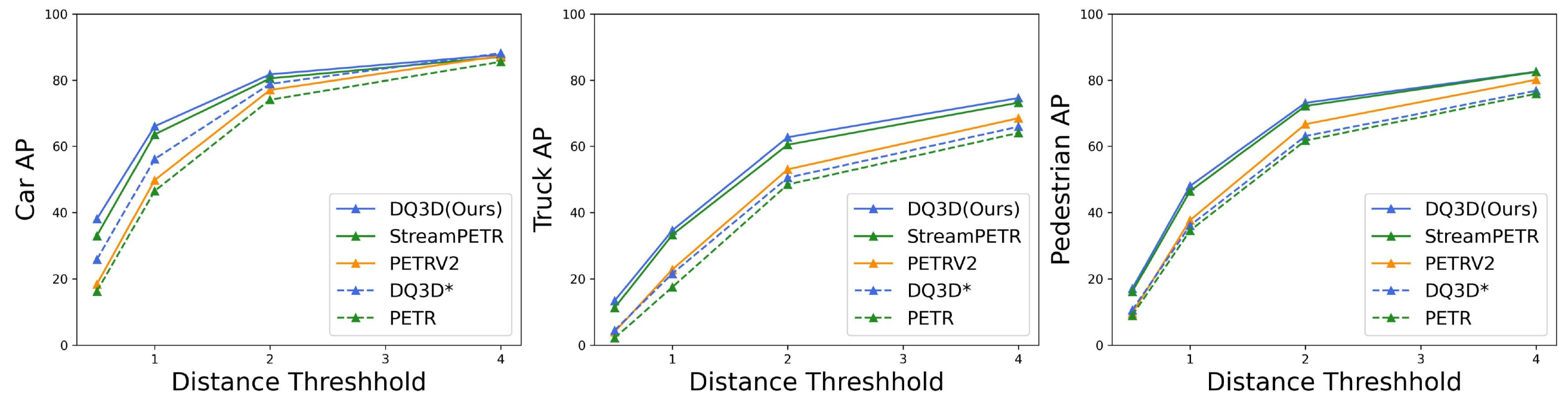}
    \caption{The mAP results with different distance thresholds on the nuScenes val set. 
    * indicates DQ3D without the Temporal Query. Solid lines represent multi-frame methods, while dashed lines denote single-frame methods.}
    \label{fig:AP}
    \vspace{-0.5cm}
\end{figure*}
We compare the DQ3D performance with state-of-the art methods on nuScenes val set and test set. The results are
shown in Table \ref{tab:performance on val} and Table \ref{tab:performance on test}.

As in the table \ref{tab:performance on val} , DQ3D with VoVNetV2 achieves 0.498 mAP and 0.582 NDS, which outperforms other methods by 1.6\% mAP and 1.1\% NDS with the same backbone and the same input image size.
From the table \ref{tab:performance on test}, DQ3D achieves 0.566 mAP and 0.625 NDS, which outperforms our baseline by 6.3\%mAP and 4.3\%NDS with the same backbone and the same input image size, delivering better performance compared to other methods.

We further visualize the detection result of our DQ3D, StreamPETR and ground truth result, as shown in Fig.\ref{fig:vis}.
The highlighted regions demonstrate how DQ3D successfully identifies objects that baseline methods either miss or misclassify. It can be observed that the baseline algorithm, StreamPETR, exhibits significant localization errors for distant objects. Additionally, the similarity between targets and the background contributes to missed detections, thereby limiting the overall performance of StreamPETR. In contrast, our enhanced method, DQ3D, demonstrates substantial improvements in detection results. The regions highlighted with red circles emphasize the more precise targets detected by our algorithm.

We compute the Average Precision (AP) according the Euclidean distance \( d \) between the 2D center points of bounding boxes in the Bird’s Eye View (BEV) perspective. This approach effectively decouples the influence of object size and orientation on AP calculations. Specifically, the distance \( d \) is set to \{0.5, 1, 2, 4\} meters. The mean Average Precision (mAP) is calculated across different object classes \( C \) and varying distance difficulties \( D \), providing a comprehensive and fair evaluation of detection performance across objects of various sizes and in diverse spatial contexts.
As illustrated in Fig.\ref{fig:AP}, we present the 3D AP results across various classifications. Our proposed DQ3D method consistently outperforms baseline methods, irrespective of whether they employ single-frame or multi-frame approaches.

\subsection{Ablation Study}

\textbf{Depth-guided queries vs. Fixed queries.}  
\begin{table}[t]
    % \centering
    \resizebox{0.5\textwidth}{!}{
    \tabcolsep=2.5pt
    \begin{tabular}{c|c c|c c|c c c c c}
    \hline
        \# & query & temp &mAP$\uparrow$& NDS$\uparrow$& mATE$\downarrow$&mASE$\downarrow$&mAOE$\downarrow$&mAVE$\downarrow$&mAAE$\downarrow$\\
        \hline
         1&fixed(500)&&0.371&0.425&0.703& 0.310& 0.483&0.888&0.232\\
         2&fixed(900)&& 0.398&0.446&0.704&0.270&0.495&0.843&0.218\\
         3&fixed(900)& \checkmark&0.482&0.571&0.610&\textbf{0.256}&\textbf{0.375}&0.263&\textbf{0.194} \\
         4&Ours &  &0.429&0.475&0.649&0.262&0.411&0.883&0.195 \\
         5&Ours & \checkmark &\textbf{0.498}&\textbf{0.582}&\textbf{0.585}&0.260&0.384&\textbf{0.240}&0.199 \\\hline
    \end{tabular}
    }
    \caption{Ablation studies for dynamic query generation on nuScenes val set. }
    \label{tab:ablation}
    \vspace{-0.5cm}
\end{table}
We first compare the detection performance using our depth-quided object queries with fixed object queries. From \#1 and \#2 in Table \ref{tab:ablation}, it can be observed that
in the case of fixed queries, a larger query amount can improve performance since the fixed query based methods rely on densely placed object queries to localize objects. 

When using multi-frame input, version \#3 outperforms the single-frame version \#2 by 8.4\% in mAP and 12.5\% in NDS, and the metric mAVE is significantly lower. This demonstrates that the temporal query can enhance the performance of 3D object detection.
According to \#5 and \#3, when replacing the fixed queries with dynamically generated queries, mAP and NDS improve by 1.6\% and 1.1\%, respectively. For a fair comparison, in the single-frame setting, mAP and NDS improve by 3.1\% and 2.9\%, respectively, as shown by \#4 and \#2.
This result demonstrate the effectiveness of our proposed depth-guided query can improve 3D detection performance on both multi-frame setting and single-frame setting. The result on multi-frame can improve that our proposed hybrid attention layer can fuse the temporal query and depth-guided query well.

\begin{figure}[t]
    \centering
    \includegraphics[width=\linewidth]{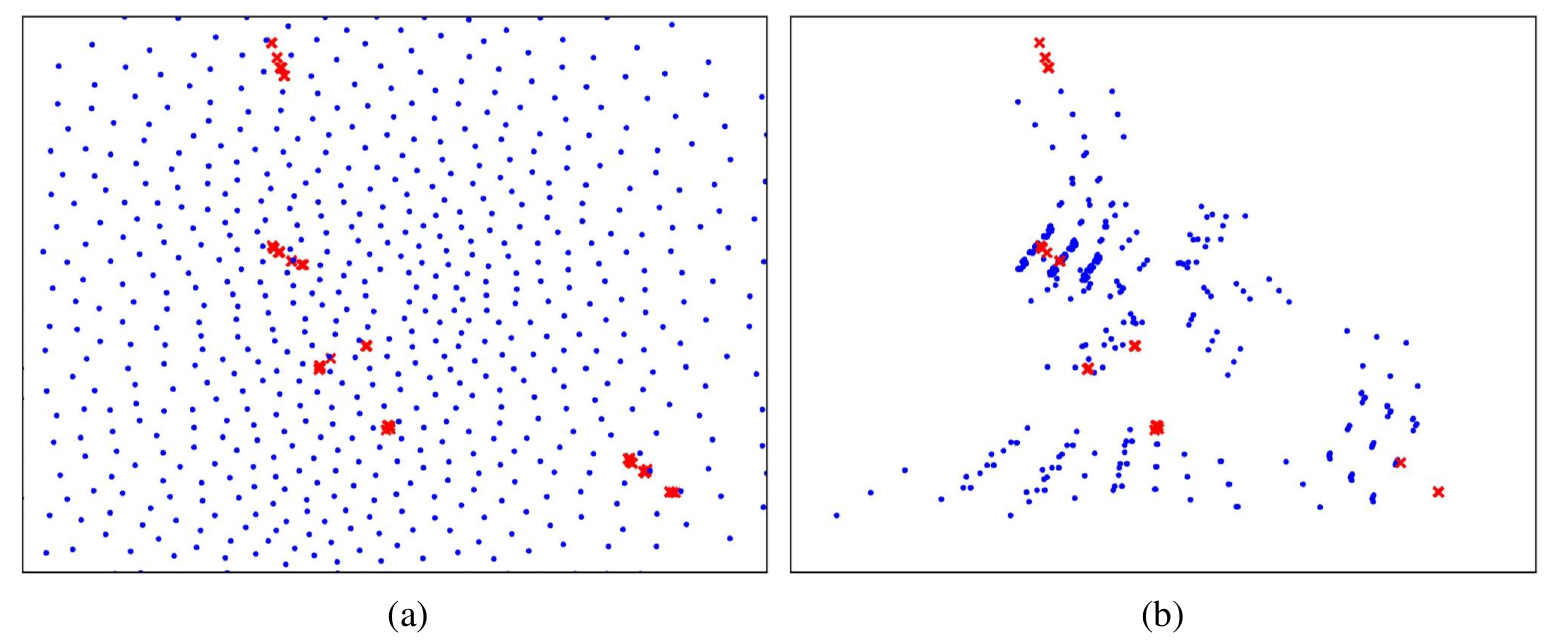}
    \caption{Visualization of normlized reference queries' Bird-Eye View(BEV). (a) only fixed query (b) depth-guided query. Red 'x' denotes ground truth center point and blue point denotes to the reference point. }
    \label{fig:ref}
    \vspace{-0.4cm}
\end{figure}

We further visualize the distribution of reference points, as illustrated in Fig.\ref{fig:ref}. Notably, unlike previous methods, the queries generated by our Depth-Guided Query (DQG) module are concentrated around the target objects. This targeted distribution enhances the model's focus on relevant regions, thereby improving detection accuracy 
by minimizing the processing of irrelevant areas.

\textbf{Ablation on different 2D Detector.}
\begin{table}[!t]
    \centering
    
    \resizebox{0.5\textwidth}{!}{
    
    \tabcolsep=2pt
    
    \begin{tabular}{c|c c|c c c c c}
    \hline
         Ours& mAP$\uparrow$& NDS$\uparrow$& mATE$\downarrow$&mASE$\downarrow$&mAOE$\downarrow$&mAVE$\downarrow$&mAAE$\downarrow$\\
    \hline
         + Faster RCNN\cite{ren2016faster}  &\textbf{0.498}&\textbf{0.582}&0.585&\textbf{0.260}&\textbf{0.384}&\textbf{0.240}&0.199\\
         + YOLOX\cite{ge2021yolox} &0.483&0.573&\textbf{0.567}&0.279&0.389&0.265&\textbf{0.186}\\
    \hline
    \end{tabular}
    }
    \caption{Comparison on different 2D Detector on nuScenes val set.}
    \label{tab:2d detector}
    \vspace{-0.5cm}
\end{table}
To demonstrate the versatility of our method with respect to 2D detectors, we incorporate several 2D detectors from the MMDetection framework in our experiments. 
The results are presented in Table \ref{tab:2d detector}. 
As shown, while the accuracy of 2D object detection does influence the performance of our 3D object detection, our method does not rely on any specific 2D detection network, highlighting its strong scalability.

\section{Conclusion}
In this paper, we propose a depth-guided 3D object query for 3D object detection(DQ3D).
In our framework, 
we utilize 2D detections and estimated depth map to sample 3D reference points 
for 3D object query generation.
We also integrate the temporal query generation method with a temporal query alignment module and a hybrid attention layer. 
In our experiments, we demonstrate promising results on nuScenes dataset with our
proposed DQ3D framework, which outperform our baseline StreamPETR on the same experiment settings in terms of the mAP(6.3\%) and NDS(4.3\%).

% \textbf{Limitations} 
% Since DQ3D generates object queries
% from depth estimation and 2D detections, 
% an object might be missed if the 2D detector fails to detect it in all the camera views.
% The inaccuracy of depth estimation also affects the performance.
% \reftitle{References}
% \section*{reference}



% \externalbibliography{yes}
\bibliographystyle{IEEEbib}
\bibliography{ref}
\end{document}